\documentclass[journal]{IEEEtran}
\usepackage{amsmath,amsfonts}
\usepackage{algorithmic}
\usepackage{algorithm}
\usepackage{array}
\usepackage[caption=false,font=normalsize,labelfont=sf,textfont=sf]{subfig}
\usepackage{textcomp}
\usepackage{stfloats}
\usepackage{url}
\usepackage{verbatim}
\usepackage{graphicx}
\usepackage{cite}

\usepackage{booktabs}
\usepackage{multirow}
\usepackage{xcolor}
\usepackage{makecell}
\usepackage[mathlines,switch]{lineno} 

\begin{document}

\title{Event-based Facial Keypoint Alignment via Cross-Modal Fusion Attention and Self-Supervised Multi-Event Representation Learning}

\author{Donghwa Kang, Junho Kim, and Dongwoo Kang,~\IEEEmembership{Member,~IEEE,}
\thanks{This work was supported in part by the Institute of Information \& Communications Technology Planning \& Evaluation(IITP) grant funded by the Korea government(MSIT) (No. RS-2024-00337012) \textit{(Corresponding author: Dongwoo Kang.)} }
\thanks{Donghwa Kang, Junho Kim, and Dongwoo Kang are with the School of Electronic and Electrical  Engineering, Hongik University, Seoul 04066, South Korea (e-mail: donghwa@g.hongik.ac.kr; darkenergy814@g.hongik.ac.kr; dkang@hongik.ac.kr)}
}

\markboth{Journal of \LaTeX\ Class Files,~Vol.~14, No.~8, August~2021}%
{Shell \MakeLowercase{\textit{et al.}}: A Sample Article Using IEEEtran.cls for IEEE Journals}

\IEEEoverridecommandlockouts
\IEEEpubid{\parbox{\textwidth}{\centering Copyright \copyright~2026 IEEE. Personal use of this material is permitted. However, permission to use this material for any other purposes must be obtained from the IEEE by sending an email to pubs-permissions@ieee.org.}}

\maketitle

\begin{abstract}
Event cameras offer unique advantages for facial keypoint alignment under challenging conditions, such as low light and rapid motion, due to their high temporal resolution and robustness to varying illumination. However, existing RGB facial keypoint alignment methods do not perform well on event data, and training solely on event data often leads to suboptimal performance because of its limited spatial information. Moreover, the lack of comprehensive labeled event datasets further hinders progress in this area. To address these issues, we propose a novel framework based on cross-modal fusion attention (CMFA) and self-supervised multi-event representation learning (SSMER) for event-based facial keypoint alignment. Our framework employs CMFA to integrate corresponding RGB data, guiding the model to extract robust facial features from event input images. In parallel, SSMER enables effective feature learning from unlabeled event data, overcoming spatial limitations. Extensive experiments on our real‑event E‑SIE dataset and a synthetic‑event version of the public WFLW‑V benchmark show that our approach consistently surpasses state‑of‑the‑art methods across multiple evaluation metrics.
\end{abstract}

\IEEEpubidadjcol
\section{Introduction}
\IEEEPARstart{F}{acial} keypoint alignment is a fundamental task in computer vision that involves accurately detecting and localizing facial landmarks\cite{edit_face1,edit_face2}. This capability is essential for applications such as face recognition\cite{faceRecog1, faceRecog2, facemask24}, human emotion analysis\cite{emotion, face-surv24}, augmented reality\cite{Pei2024CVPR}, advanced driver assistance systems\cite{driver_system}, surveillance\cite{faceattack25}, and human-robot interaction\cite{ref-ral1}. Although RGB-based methods have achieved impressive results\cite{slpt22,star23}, they often struggle in dynamic scenarios. Factors such as low illumination and rapid movements contribute to less reliable keypoint detection, indicating that the problem remains challenging\cite{challenge1, challenge2}. These issues have led to growing interest in event cameras\cite{event_camera}.

Event cameras are next-generation, neuromorphic image sensors that detect brightness changes asynchronously at the pixel level\cite{camera_sensor}. Unlike conventional RGB cameras that capture frames at fixed intervals, each pixel in an event camera records an event only when the change in brightness exceeds a threshold\cite{event_TETCI}. These sensors offer advantages such as low power consumption, high dynamic range, and high temporal resolution. Owing to these strengths, event cameras have emerged as a promising tool in various vision systems, with applications in event-based classification\cite{eventClass1,edit_event1}, object recognition\cite{eventObjectRecognition}, and depth and optical flow estimation\cite{opticalflow}. Despite these advantages, event data tend to be noisy and lack rich visual information, such as texture, color, and contextual appearance. Consequently, many studies are focusing on overcoming these challenges\cite{eventObjectRecognition, leod, eventdance}.
\IEEEpubidadjcol

\begin{figure}[t]
  \centering
   \includegraphics[width=1\linewidth]{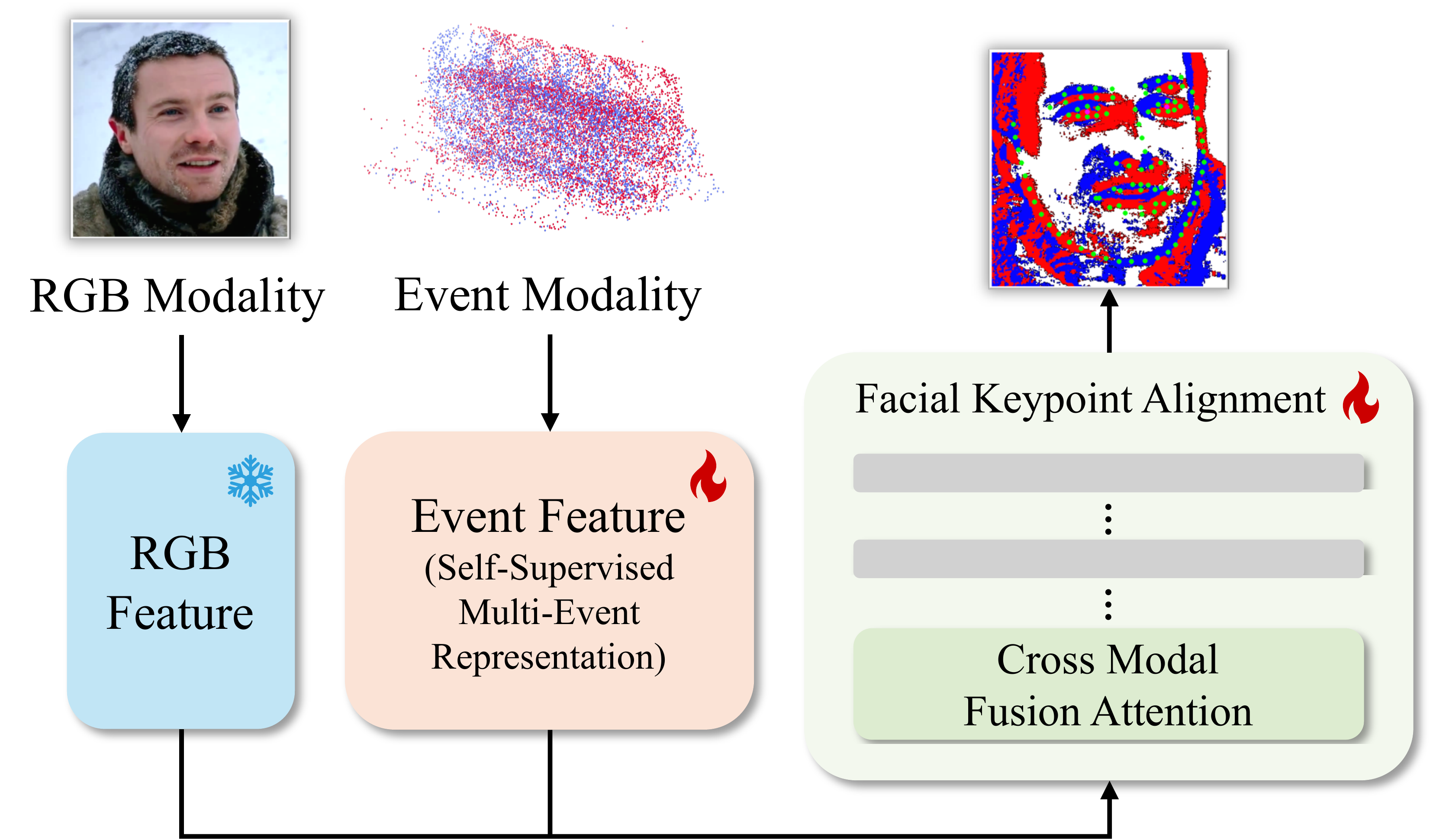}
   \caption{Our framework combines the rich spatial detail of RGB frames with the robust motion cues of event data. We train the event backbone through self-supervised multi-event representation learning and integrate both modalities with cross-modal fusion attention in a transformer-based pipeline.}
   \label{fig:concept}
\end{figure}

Compared to the above event-based vision tasks, event-based facial keypoint alignment remains in its early stages of research\cite{event_face, event_motion}. In contrast to the abundance of public RGB face datasets\cite{celebvhq, celeba}, event camera face datasets are limited, and capturing and precisely labeling keypoints is more challenging due to the inherent lack of spatial information in event data\cite{nefer23}. Additionally, directly applying existing RGB-based models to event data results in poor performance because of the significant domain gap between the two modalities. A network model and training strategy that considers the unique characteristics of event face data is desirable to replace conventional supervised learning approaches used in RGB facial keypoint alignment\cite{daehyun23}.

To address the issues above, we propose an event-based facial keypoint alignment framework that utilizes cross-modal fusion attention (CMFA) combined with self-supervised multi-event representation learning (SSMER). As illustrated in Fig. \ref{fig:concept}, our framework incorporates CMFA into a transformer-based facial alignment network, allowing the network to exploit the rich texture information from RGB data to compensate for the limitations of event data and achieve accurate landmark alignment. In addition, we replace the conventional backbone with SSMER module that enables the event feature extractor to be trained without relying on labeled event facial keypoint datasets. Specifically, our module fuses three representations, namely frame\cite{frame}, voxel\cite{voxel18} and timesurface\cite{ts17}, improving fine-tuning performance compared to typical self-supervised learning (SSL) methods\cite{simclr} that utilize only one modality.

To further address the scarcity of event face training data, our framework generates synthetic event face data, E-CelebV-HQ, from an existing public RGB dataset CelebV-HQ\cite{celebvhq}. Finally, to enable accurate and comprehensive evaluation on real event cameras, we introduce E-SIE, a novel, self-collected real event-based facial dataset capturing Speed, Illumination, and Eyeglasses variations. E-SIE includes normal and low illumination, varied motion speeds, diverse head poses, and the presence or absence of eyeglasses, with synchronized RGB and event data to facilitate reliable benchmarking for event-based facial analysis. In summary, the contributions of this paper are as follows:

\begin{itemize}
    \item We present an event-based facial keypoint alignment framework with CMFA that integrates the complementary strengths of event data and RGB for improved performance.

    \item To address the scarcity of labeled event face datasets, we propose an SSMER-based event backbone, enabling diverse representations from unlabeled event data.

    \item We propose E‑CelebV‑HQ, a large synthetic‑event training set spanning 83 facial attributes whose diversity enables our backbone to learn and generalize without manual labels.
    
    \item We introduce E-SIE, a real event-based facial dataset capturing speed, illumination, and eyeglasses variations to validate our algorithm.
\end{itemize}

\section{Related Work}
\subsection{RGB Camera-based Facial Keypoint Alignment}
With the advent of deep learning, coordinate regression methods for directly mapping facial landmark coordinates emerged.\cite{mdm16} employs a convolutional recurrent network for cascaded regression, while\cite{sdfl21} uses graph convolutional networks to refine landmark relationships.\cite{wing18} introduces a loss emphasizing small and medium-range errors.\cite{slpt22, repformer22} propose transformer-based architectures to capture global facial structure.\cite{slpt22} extracts local patches for each landmark and aggregates them using an adaptive attention-based mechanism.

In parallel, heatmap-based approaches generate a two-dimensional probability map for each landmark by selecting the highest-intensity pixel as its location. Techniques such as\cite{lab18} regress facial boundary information to improve accuracy under challenging conditions.\cite{awing19} focuses on reducing small errors near key landmark locations, producing robust heatmap predictions.\cite{hrnet21} utilizes parallel multi-scale feature fusion and repeated information exchange to effectively improve spatial precision, and \cite{adnet21, star23} utilize anisotropic constraints and distributions to reduce semantic ambiguity. Hybrid methods combine coordinate regression and heatmap generation, with \cite{pipnet21} providing coarse heatmaps and offset predictions simultaneously for efficient landmark localization.

\begin{figure*}
  \centering
  \includegraphics[width=0.99\linewidth]{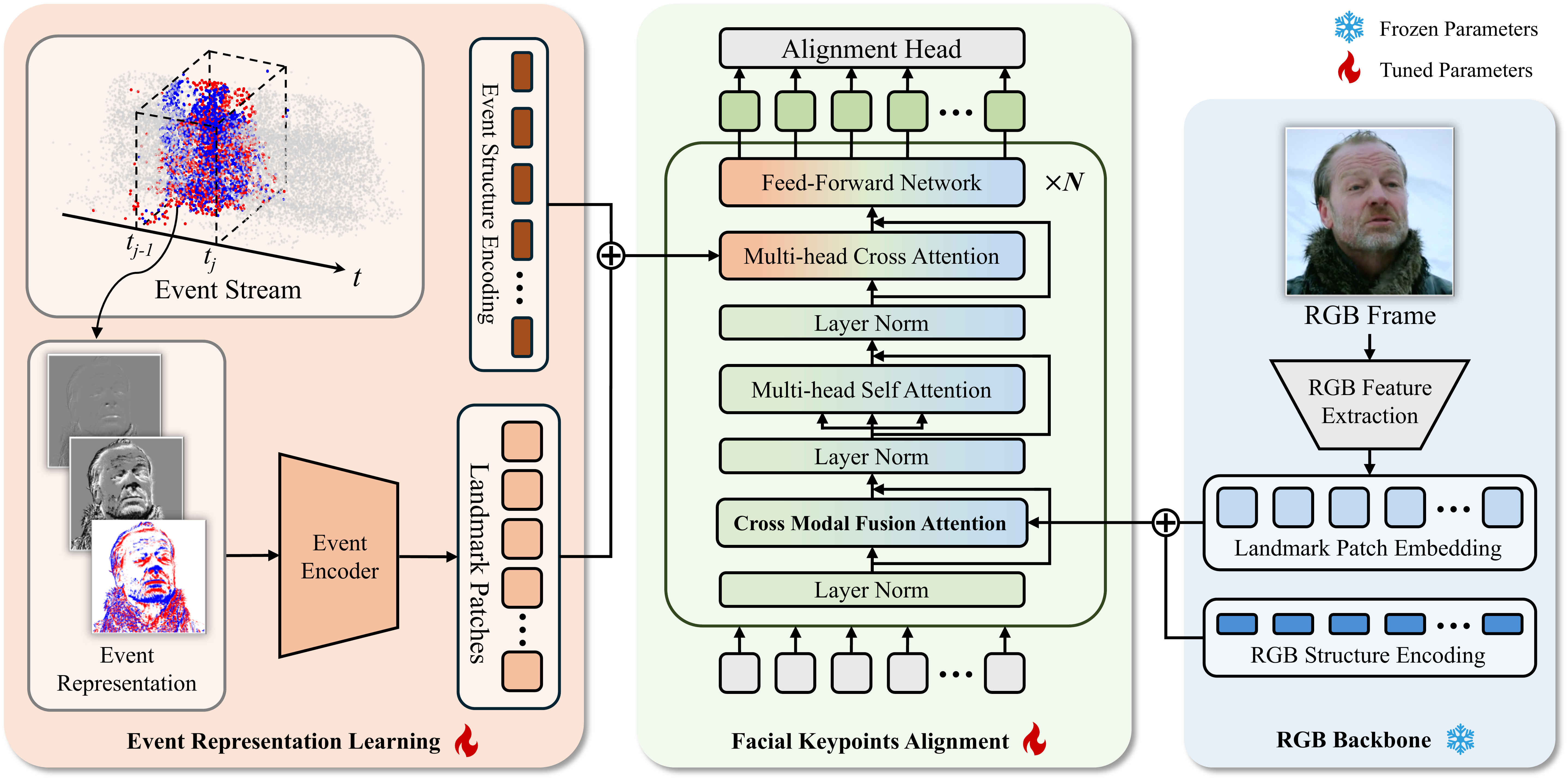}
  \caption{An overview of our event-based facial keypoint alignment pipeline. The synchronized RGB and event streams each pass through feature extraction: the RGB branch employs a pretrained backbone, while the event branch uses our SSMER backbone. The extracted features, together with their respective structure encodings, are then fused by CMFA, followed by MSA and MCA to refine landmark-specific patches. A shared alignment head subsequently predicts the facial landmarks.}
  \label{fig:network}
\end{figure*}

\subsection{Self-Supervised Learning for Pretraining}
SSL trains a network without labeled data, enabling learned representations to be applied to downstream tasks such as classification and detection\cite{ssl_class1, ssl_class2}. The joint-embedding framework, a variant of SSL, learns features by minimizing distances between positive pairs and maximizing distances between negative pairs. These frameworks divide into contrastive and non-contrastive methods. Contrastive approaches minimize positive-pair distances while maximizing negative-pair distances, as in SimCLR\cite{simclr}. In contrast, non-contrastive methods focus solely on reducing positive-pair distances. Although non-contrastive methods are computationally efficient, they risk output collapse with symmetric architectures. To mitigate this risk, some incorporate a predictor module\cite{simsiam,byol}, use a momentum encoder\cite{byol}, or employ a clustering step\cite{swav}.

SSL has also been explored in event-based tasks such as object classification\cite{objectClass}, object recognition\cite{eventPretraining}, and object keypoint detection\cite{sd2event}. However, its application to facial keypoint alignment has yet to be investigated. In this work, we propose a modified version of SimSiam\cite{simsiam} for SSL-based multi-event representation learning, serving as the backbone for event-based facial keypoint alignment.

\subsection{Event-Based Vision for Facial Analysis}
Event-based vision has been extensively studied for object-related tasks such as classification and recognition using temporal cues\cite{eventdance, leod, edit_event2}. In facial analysis, event cameras have been applied to face detection, pose alignment, expression and emotion recognition, and gaze tracking, although these applications remain in early stages compared to object-based tasks. For example,\cite{wacv16} presented a patch-based model demonstrating high-speed, power-efficient performance,\cite{ryan21} introduced a recurrent YOLO for face detection. In face pose alignment,\cite{savran-sensors20} used a cascaded regression of tree ensembles while\cite{savran-cviu23} proposed a framework that adaptively generates sparse pose-events at variable rates using detection-based timescale selection. Event cameras have also been employed for expression recognition\cite{micro22, nefer23, emotion-eccv24}, and for gaze analysis, with systems achieving ultra-fast tracking\cite{10khz21, gaze23, mambapupil24}. 

Research on event-based facial keypoint alignment remains limited. Recently, some works have introduced synthetic data and advanced frameworks for event-based processing. For instance,\cite{daehyun23} proposed an adaptive styleflow approach to generate synthetic event data as a training set, along with a cross-modal learning strategy combining RGB and event modalities. Meanwhile,\cite{neurodfd22} developed a framework that learns spatio-temporal characteristics using a feature pyramid shift network and a context shift module, manually labeling a real event dataset. However, the limited spatial detail of event data often constrains performance, especially for precise tasks like facial keypoint alignment.

\begin{figure}
  \centering
   \includegraphics[width=0.9\linewidth]{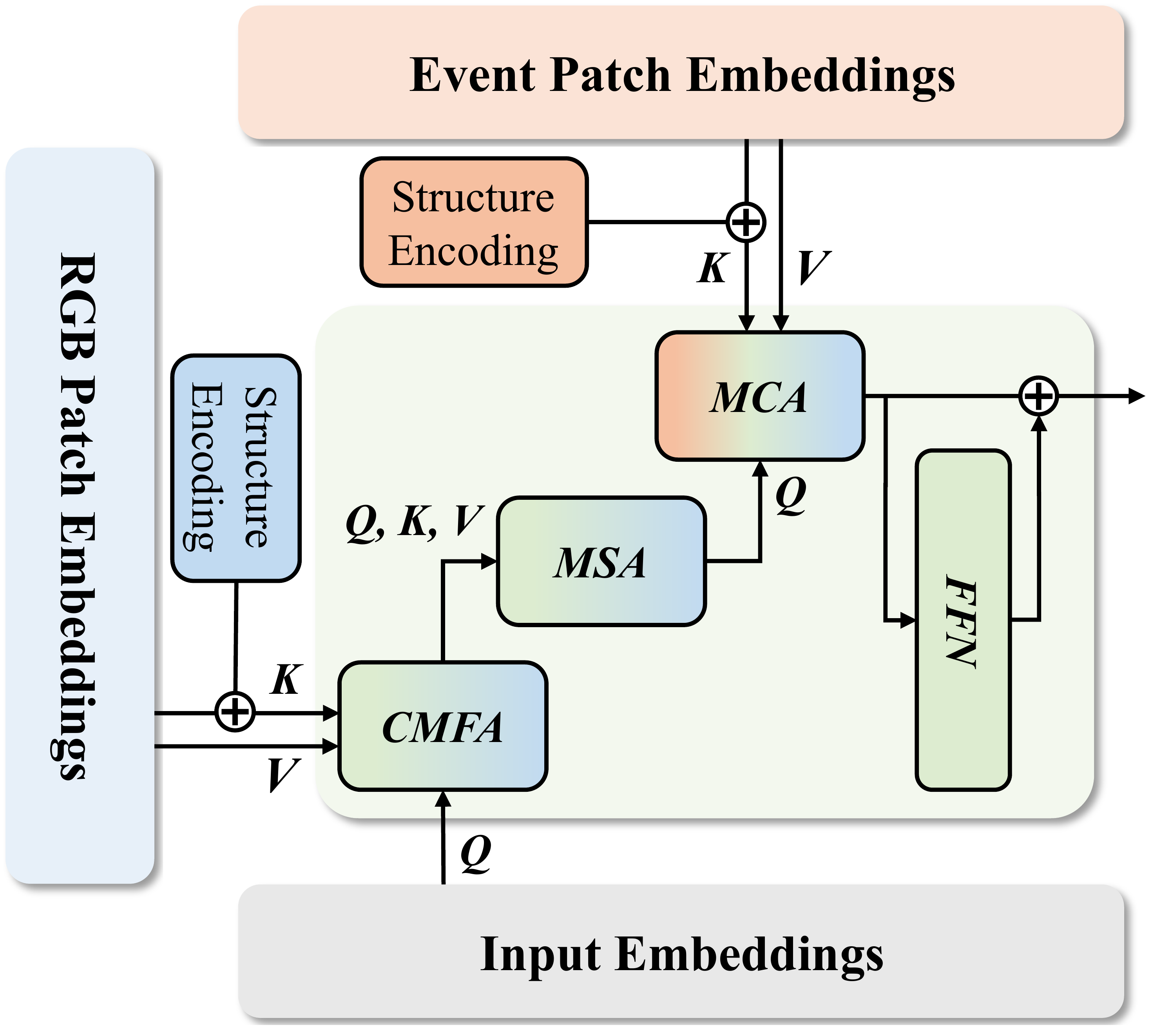}

   \caption{Detailed layout of our facial keypoint alignment module, illustrating how CMFA works in with MSA and MCA. CMFA receives the query from the input embeddings and obtains the key and value from RGB patch embeddings with structure encoding, fusing them. The resulting refined features then flow through MSA and MCA, where event patch embeddings with structure encoding further guide the final keypoint alignment.}
   \label{fig:cmfa}
\end{figure}

\section{PROPOSED METHOD}
\subsection{Overview}
\label{sub:overview}
Facial keypoint alignment in event camera data poses a distinct challenge. While event cameras excel in low-light and rapid-motion conditions, they lack the spatial detail found in RGB data. Our approach merges these complementary strengths and limitations by integrating both RGB and event modalities in a unified framework. As illustrated in Fig. \ref{fig:network}, our CMFA and SSMER-based event facial keypoint alignment pipeline processes synchronized RGB images and event data in parallel. The RGB images pass through a pretrained backbone built upon SLPT\cite{slpt22} pretrained on WFLW\cite{lab18}, to extract features. Meanwhile, the event data is transformed into selected event frame representations and passed through our proposed SSMER event backbone, which employs multiple event representations and contrastive learning to enhance feature generalization and training stability. Within the facial keypoint alignment flow, we introduce CMFA to use the richer RGB features as a key for refining the more fluctuating event features.

Our approach proceeds in two distinct stages. First, we train the event backbone through a self-supervised method using synthetic data generated by an event simulator\cite{v2e}, producing E-CelebV-HQ. This large-scale synthetic dataset provides the diversity needed for robust SSL training of the event backbone. In the second stage, we conduct supervised fine-tuning with our CMFA block for event-based facial keypoint alignment. We derive pseudo labels by applying an RGB-based facial keypoint alignment method\cite{slpt22} to the original RGB data\cite{celebvhq}, treating these labels as ground truth for the event-based model. In Sec.\ref{sub:cmfa}, we detail the CMFA block, which uses RGB features to refine event representations for more stable feature learning. Next, Sec.\ref{sub:ssmr} explains our SSMER backbone, incorporating multiple event representations and contrastive learning to enhance generalization and training stability. Finally, in Sec.\ref{sub:E_SIE}, we introduce E-SIE, our self-collected real event face dataset for thorough evaluation.

\subsection{Cross-Modal Fusion Attention}
\label{sub:cmfa}

The CMFA block refines event features by incorporating complementary RGB information, enabling more stable landmark estimation. Inspired by SLPT\cite{slpt22}, we present a Transformer-based facial keypoint alignment architecture enhanced with CMFA, which utilizes RGB features and structure encoding as a query to guide event-based predictions. As illustrated in Fig. \ref{fig:network}, the embedding features extracted from the pretrained event backbone and the pretrained RGB backbone successively pass through CMFA, multi-head self-attention (MSA), and multi-head cross-attention (MCA), and are then fine-tuned for keypoint prediction in a supervised manner.

Conventional vision transformers typically use multi-head self-attention (MSA). Existing transformer-based facial keypoint alignment methods\cite{slpt22} add multi-head cross-attention (MCA) with structure encoding to capture landmark relationships. Our approach extends this idea by introducing CMFA, creating three sequential blocks CMFA, MSA, and MCA. As shown in Fig. \ref{fig:cmfa}, the input embeddings first enter CMFA, where the RGB patch embeddings combined with structure encoding act as the key, while the original RGB features serve as the value. The refined outputs then pass through MSA and eventually proceed to MCA, which applies event patch embeddings with structure encoding as the key for final facial keypoint alignment. By placing CMFA at the beginning, event features can immediately benefit from the spatial details of RGB, resulting in more stable learning under challenging conditions.

\begin{figure}
  \centering
   \includegraphics[width=1\linewidth]{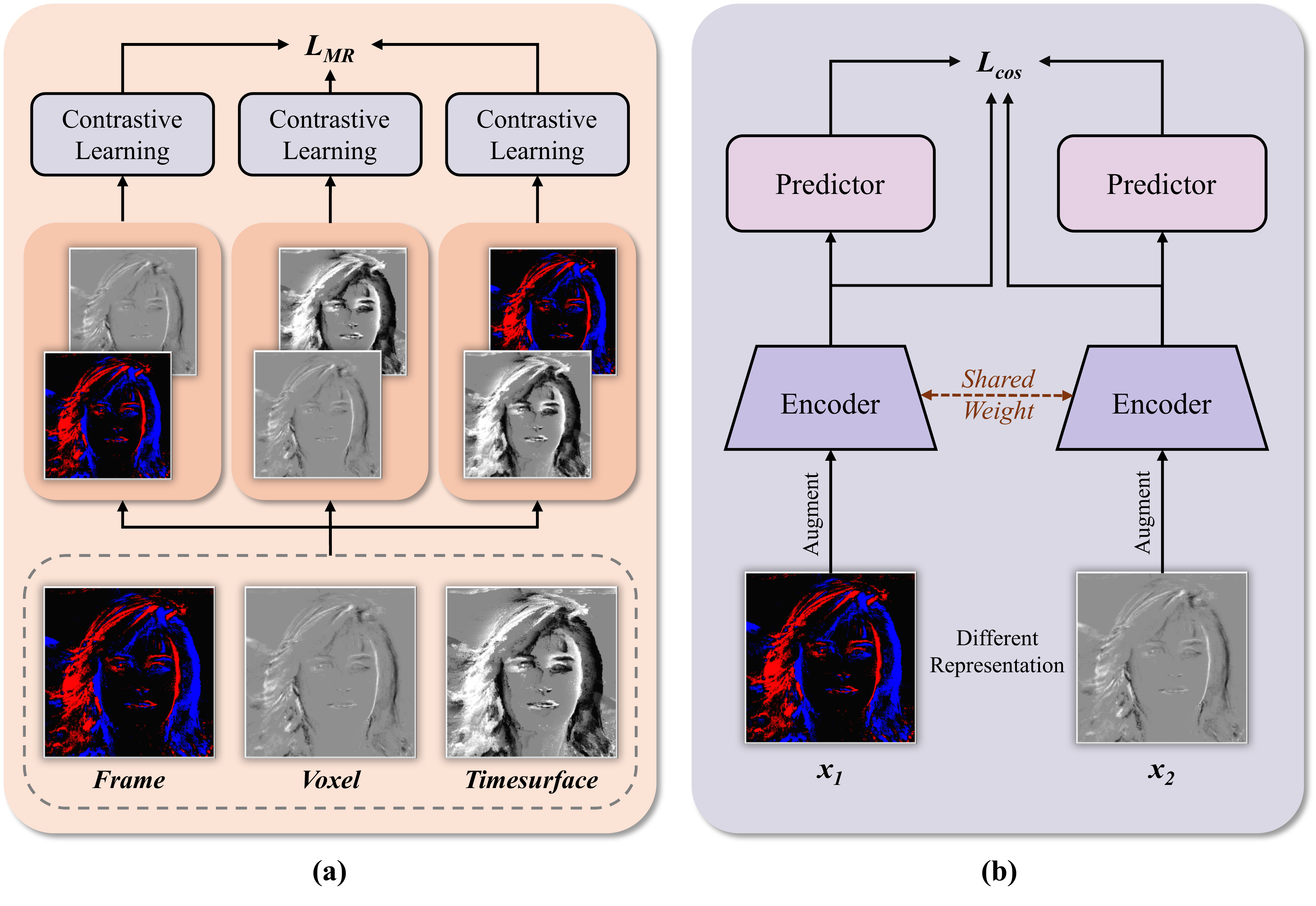}
   \caption{Illustration of our SSMER pipeline. (a) Three representation pairs are processed via contrastive learning, and the resulting losses are summed into a multi-representation loss. (b) A detailed view of our contrastive learning procedure for each pair.}
   \label{fig:ssmer}
\end{figure}

To formally define CMFA, let $\boldsymbol{T}^l$ be the input embedding for the $l$-th layer, and $\boldsymbol{Q}$ represents the landmark query. We set the pretrained RGB features $\boldsymbol{F}_h^{rgb}$  and their corresponding structure encoding $\boldsymbol{P}_h^{rgb}$. The attention weights $\boldsymbol{A}_h$ in the $h$-th head of CMFA are obtained by

\begin{equation}
 \resizebox{\columnwidth}{!}{$
  \boldsymbol{A}_h=softmax \left( \frac{(\boldsymbol{T}_h^{l} + \boldsymbol{Q}_h)\boldsymbol{W}_h^q((\boldsymbol{F}_h^{rgb}+\boldsymbol{P}_h^{rgb})\boldsymbol{W}_h^k)^T}{\sqrt{C_h}} \right),$}
  \label{eq:cmfa}
\end{equation}

where $C_h$ is the feature dimension per head, and $\boldsymbol{W}_h^q$ and $\boldsymbol{W}_h^k$ are learnable parameters. The CMFA output for all $\textit{H}$ heads at the $l$-th layer is then computed as

\begin{equation}
   CMFA(\boldsymbol{T}^l) = \left[ \boldsymbol{A}_1 \boldsymbol{T}_1^l \boldsymbol{W}_1^v ; \dots ; \boldsymbol{A}_H \boldsymbol{T}_H^l \boldsymbol{W}_H^v \right] \boldsymbol{W}_P,
  \label{eq:total}
\end{equation}

with $\boldsymbol{W}_h^v$  and $\boldsymbol{W}_P$  being additional learnable parameters. We also apply a residual connection and layer normalization $LN$ as

\begin{equation}
  \boldsymbol{T}^{(l)} = \boldsymbol{T}^{(l-1)} + CMFA \left( LN \left( \boldsymbol{T}^{(l-1)} \right) \right).
  \label{eq:cmfa-residual}
\end{equation}

Finally, the embeddings updated by CMFA pass through MSA and MCA, producing $\boldsymbol{T}^{'(l)}$, the final output of the $l$-th layer, which is calculated by

\begin{equation}
  \begin{aligned}
    \boldsymbol{T}'^{(l)} &= \boldsymbol{T}^{(l)} + MSA \left( LN \left( \boldsymbol{T}^{(l)} \right) \right) \\
    &\quad + MCA \left( LN \left( \boldsymbol{T}^{(l)} + MSA \left( LN \left( \boldsymbol{T}^{(l)} \right) \right) \right) \right).
  \end{aligned}
  \label{eq:msa-mca}
\end{equation}

\noindent This sequential application of CMFA, MSA, and MCA fuses spatially rich RGB information with event embeddings, yielding robust facial keypoint alignment.

During the fine-tuning stage, the refined features from the CMFA module are passed through the alignment head, which performs the final landmark coordinate regression using the Normalized L2 loss. The objective function is defined as:

\begin{equation}
\mathcal{L}_{reg} = \frac{1}{S \cdot N \cdot K} \sum_{i=1}^{S} \sum_{j=1}^{N} \sum_{k=1}^{K} \frac{\| \mathbf{y}_{gt}^{(k)} - \mathbf{y}_{pred}^{(i,j,k)} \|_2}{d}
\label{eq:cmfa-loss}
\end{equation}

\noindent{where $S$, $N$, and $K$ denote the number of coarse-to-fine stages, the number of inherent relation layers, and the total number of landmarks, respectively. The indices $i, j,$ and $k$ correspond to the $i$-th stage, $j$-th layer, and $k$-th landmark. $\mathbf{y}_{gt}^{(k)}$ represents the ground-truth coordinate vector of the $k$-th landmark, while $\mathbf{y}_{pred}^{(i,j,k)}$ indicates the predicted coordinate vector derived from the $j$-th layer in the $i$-th stage.}

\subsection{Self-Supervised Multi-Event Representation Learning}
\label{sub:ssmr}

\begin{table*}
  \centering
    \caption{Characteristics of the Event Face Dataset.}
    \resizebox{1\textwidth}{!}{
  \begin{tabular}{lccccccc}
      \toprule
Dataset& RGB& \makecell{RGB-Event \\Sync}& \# of Videos& \makecell{\# of \\Keypoints}&\makecell{Event Data\\(Real/Synthetic)}&Subjects &Sensor\\
  \midrule
    N-Caltech 101\cite{ncaltech15}& Caltech 101\cite{caltaech}& -& 435&  -&Real & 385&ATIS\cite{ref-atis}\\
    NEFER\cite{nefer23}& \checkmark& -& 597& -&Real & 29&Gen4 CD\cite{ref-gen4}\\
 FES\cite{fes24}& -& -& 3,780& 5 & Real & 18&Gen3 ATIS\cite{ref-gen3atis}\\
  \midrule
    \textbf{E-CelebV-HQ (ours)}& CelebV-HQ\cite{celebvhq}& \checkmark& 35,664& 98 &Synthetic & 15,653&-\\
 \textbf{E-WFLW-V(ours)}& WFLW-V\cite{wflwv}& \checkmark& 1,000& 98& Synthetic& -&-\\
    \textbf{E-SIE (ours)}& \checkmark& \checkmark& 72& 5 &Real & 9&DAVIS346\cite{ref-davis346}\\
    \bottomrule
  \end{tabular}
    }
  \label{tab:dataset}
\end{table*}

Event-based facial keypoint datasets are limited, and event data inherently lack the spatial detail found in RGB images. To address these challenges, we propose a self-supervised multi-event representation learning approach for the event backbone. Rather than relying on a single event representation, our multi-representation strategy integrates multiple forms of event data, enabling more diverse feature extraction and producing a more robust backbone despite the limited spatial cues in raw event data. 

Our SSMER combines three event representations, frame\cite{frame}, voxel\cite{voxel18}, and timesurface\cite{ts17}, to provide a richer feature set than a single representation can offer. As illustrated in Fig. \ref{fig:ssmer} (a), we form three pairs of representations frame-voxel, voxel-timesurface, and timesurface-frame, each passed through a SimSiam-based\cite{simsiam} contrastive pipeline. Given two augmented inputs $x_1$ and $x_2$, we encode them into $z_1$ and $z_2$ and produce $p_1$ and $p_2$ via a predictor. Formally, we use $f$ as the encoder and $h$ as the predictor: $z_1 = f(x_1)$, $z_2 = f(x_2)$ and $p_1 = h(z_1)$, $p_2 = h(z_2)$. We measure the distance $\mathcal{D}(p_1, z_2)$ as 

\begin{equation}
\mathcal{D}(p_1, z_2) = -\frac{p_1}{{\|p_1\|}_2}\cdot\frac{z_2}{{\|z_2\|}_2}.
\label{eq:dist_cosine}
\end{equation}

\noindent{Following SimSiam\cite{simsiam}, we treat the target features $z$ with stop-gradient and adopt cross pairing, matching $p_1$ to $z_2$ and $p_2$ to $z_1$. This asymmetry helps prevent trivial constant solutions in Siamese training without negatives. We then compute the two-direction loss $\mathcal{L}_{\mathrm{cos}}$ for the $i$-th pair as

\begin{equation}
\mathcal{L}_{cos}^{i} = \frac{1}{2}\mathcal{D}(p_1, z_2) + \frac{1}{2}\mathcal{D}(p_2, z_1).
\label{eq:loss_sym}
\end{equation}

\noindent Summing these losses over all three representation pairs yields the multi-representation loss $\mathcal{L}_{MR}$ as

\begin{equation}
  \begin{aligned}
    \mathcal{L}_{MR} &= \sum_{i=1}^{3} \mathcal{L}_{cos}^{i}.
  \end{aligned}
  \label{eq:mr}
\end{equation}

\begin{figure}
  \centering
   \includegraphics[width=0.9\linewidth]{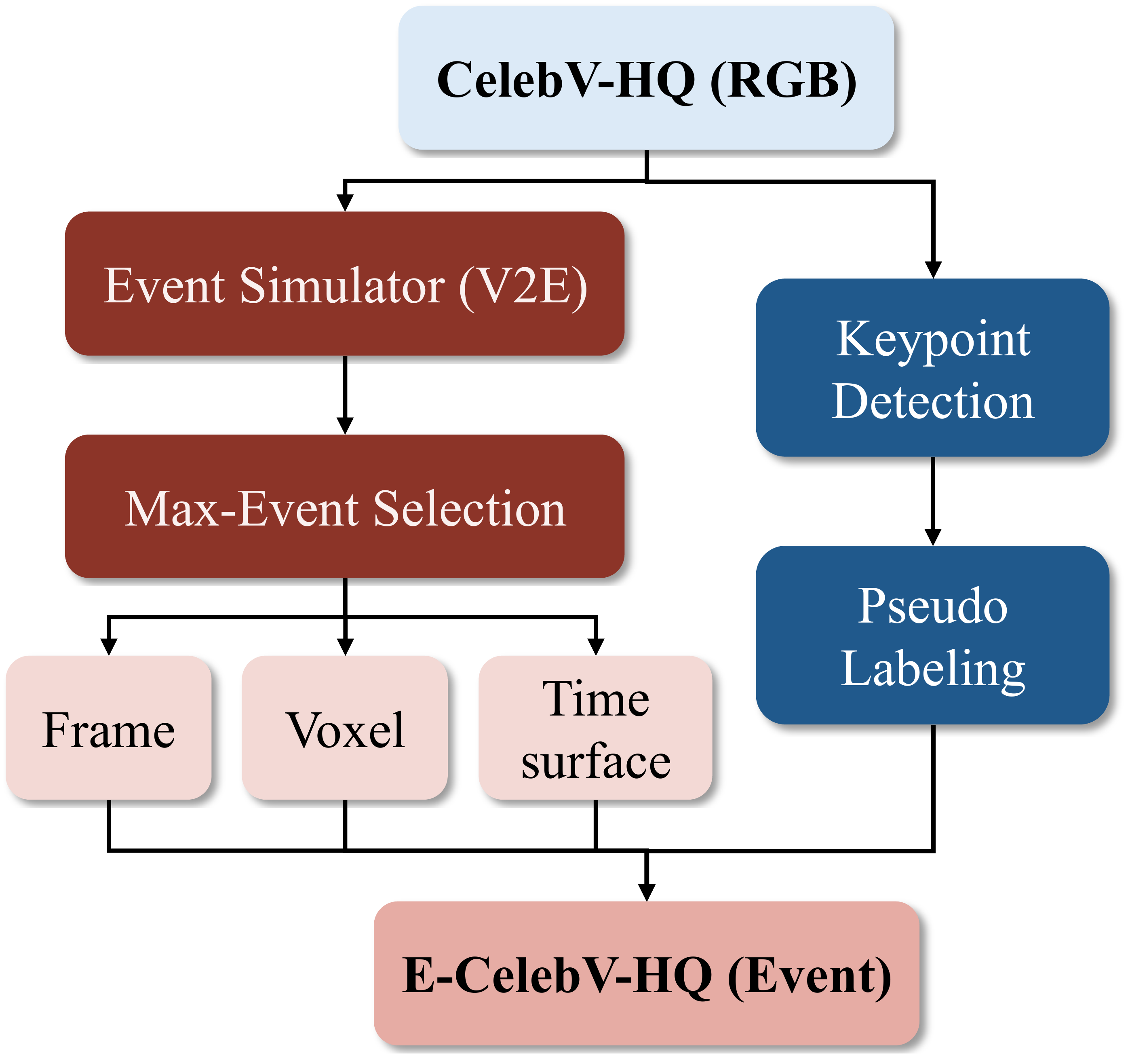}

   \caption{Construction pipeline of the synthetic E-CelebV-HQ dataset.}
   \label{fig:datapipe}
\end{figure}

Fig. \ref{fig:ssmer} (b) provides a detailed view of the contrastive learning procedure for one representation pair. We adopt HRNet\cite{hrnet21} as the backbone for feature extraction and add two layers of BatchNorm, ReLU, and fully connected layers for the projector. The predictor contains two layers, each consisting of a linear layer, BatchNorm, and ReLU. This multi-event representation SSL approach produces a more diverse and robust feature extractor for event data, compensating for the limited spatial cues inherent to event cameras.

\subsection{E-SIE: A Real Event Face Dataset}
\label{sub:E_SIE}

E-SIE is designed to capture a diverse set of facial conditions using both an event camera and an RGB camera for facial keypoint alignment. It addresses three primary variations in the order of speed, illumination, and eyeglasses. We collected approximately 10 seconds per subject using a DAVIS346\cite{ref-davis346} sensor at a resolution of 346 $\times$ 260, with a frame rate of 25 frames per second and an event bandwidth of 1,200,000 events per second. Each participant was seated about 50 cm from the camera in an indoor environment. All recordings maintain a clear view of the eyes, nose, and mouth, and partitions isolate the subject from background distractions. Because the DAVIS346 device is synchronized, both RGB and event streams share identical durations and content.

We first split each recording, which runs slightly over 10 seconds, into two halves. In each half, we select a 5-second interval in the middle and further divide it into five 1-second segments. For each of these segments, we accumulate the first 40ms of events to create a single event frame. This process yields 10 event frames in total (5 from each half). For each sampled frame, we manually annotated five keypoints in the following order: nose, left mouth corner, right mouth corner, left eye, and right eye, along with a bounding box. All human participants in this study provided their informed consent.

\section{Experiments}
\label{sec:experiments}

\subsection{Datasets}
\label{sub:dataset}

\begin{table}[t]
  \centering
  \caption{Architecture for CMFA \& SSMER Learning}
  \label{tab:param}
  \resizebox{1\columnwidth}{!}{
  \begin{tabular}{llcc}
    \toprule
    Stage & Module & Parameters (M) & Trainable \\
    \midrule
    \multirow{5}{*}{SSMER} 
    & Encoder & 3.47 & \checkmark \\
    & Projector & 9.03 & \checkmark \\
    & Predictor & 2.10 & \checkmark \\
    \cmidrule{2-4}
    & Total Params. & 14.60 & \checkmark \\
    & Trainable Params. & 14.60 & \checkmark \\
    \midrule
    \multirow{8}{*}{CMFA} 
    & RGB backbone & 6.85 & - \\
    & Event backbone & 6.82 & \checkmark \\
    & CMFA & 1.58 & \checkmark \\
    & MSA & 1.58 & \checkmark \\
    & MCA & 1.58 & \checkmark \\
    & FFN & 3.22 & \checkmark \\
    \cmidrule{2-4}
    & Total Params. & 21.62 & - \\
    & Trainable Params. & 14.78 & \checkmark \\
    \bottomrule
  \end{tabular}
  }
\end{table}

\begin{table*}
\caption{Performance comparison on event-based facial keypoint alignment, evaluated on E-SIE and E-WFLW-V \\(all models fine-tuned on E-CelebV-HQ).}
  \centering
    \resizebox{0.8\textwidth}{!}
    {
    \begin{tabular}{lccc ccc}
    \toprule
    \multirow{2}{*}{Models (RGB + Event)} & \multicolumn{3}{c}{E-SIE (Real)} & \multicolumn{3}{c}{E-WFLW-V (Synthetic)} \\
    \cmidrule(lr){2-4} \cmidrule(l){5-7}
    & NME(\%) $\downarrow$ & FR\textsubscript{10}(\%) $\downarrow$ & AUC\textsubscript{10} $\uparrow$ 
    & NME(\%) $\downarrow$ & FR\textsubscript{10}(\%) $\downarrow$ & AUC\textsubscript{10} $\uparrow$ \\
    \midrule
    HRNet\cite{hrnet21}& 28.092 & 93.861 &0.016 
& 16.718 & 56.018 &0.184 
\\
 SDFL\cite{sdfl21}& 21.304 & 74.278 & 0.100 
& 12.517 & 44.047 &0.281 
\\
 AWing\cite{awing19}& 14.754 
& 52.861 & 0.193 & 5.619 & 11.767 &0.574\\
 Wing\cite{wing18}& 12.606 & 46.000 &0.238 &    
5.313 & 10.885 &0.577 \\
 PIPNet\cite{pipnet21}& 12.488 & 60.778 & 0.118 & 15.552& 64.014 &0.146 \\
 STAR\cite{star23}& 11.793 & 45.083 & 0.236 & 5.200 & \textbf{10.540} &\textbf{0.590} \\
 ADNet\cite{adnet21}& 11.434 & 42.611 & 0.250 & 6.768 & 14.877 &0.563 \\
    \midrule
 Ours (Unfrozen RGB)& \textbf{8.090}& 26.972& 0.348& 5.081& 11.750&0.546\\
    \textbf{Ours} & 8.162 & \textbf{25.861}&\textbf{0.376}&    \textbf{4.731}& 10.623&0.579\\
    \bottomrule
  \end{tabular}
}
  \label{tab:res_esie_ewflwv}
\end{table*}

\subsection{Implementation Details}
\label{sub:implementation}

\begin{figure*}
  \centering
   \includegraphics[width=1\linewidth]{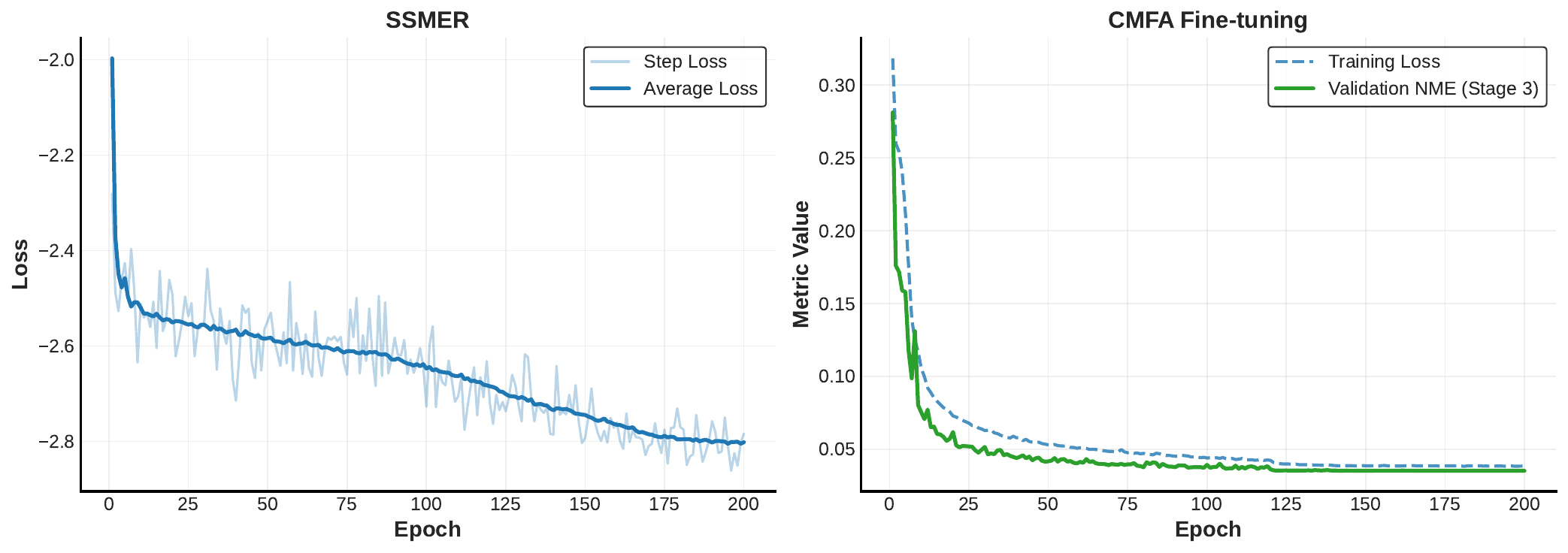}
   \caption{Training and validation responses. Left: step and average losses during SSMER pre-training. Right: training loss and validation NME during CMFA fine-tuning.}
   \label{fig:curve}
\end{figure*}

\begin{figure*}[t]
  \centering
   \includegraphics[width=1\linewidth]{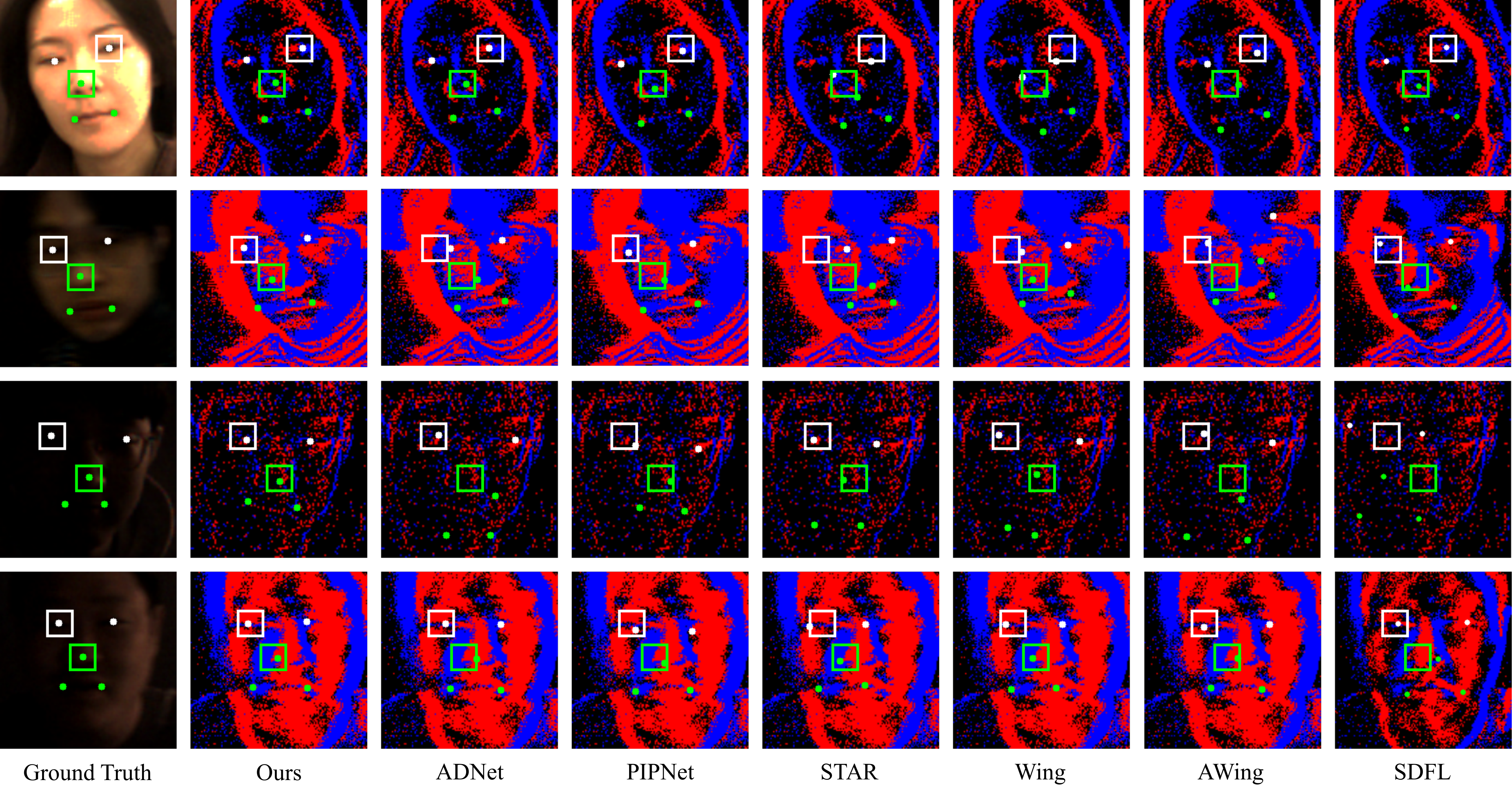}
   \caption{Qualitative comparisons with existing methods on our E-SIE dataset. Rows 1 to 4 describe normal conditions, fast speed with eyeglasses, low illumination with eyeglasses, and low illumination with fast movement, respectively.}
   \label{fig:comparison}
\end{figure*}

We aim to enhance event-based performance by using synchronized RGB and event data. As shown in Table~\ref{tab:dataset}, we explored existing face datasets, including N-Caltech 101\cite{ncaltech15}, NEFER\cite{nefer23}, and FES\cite{fes24}. The N-Caltech dataset, recorded using the ATIS sensor\cite{ref-atis}, was created by displaying images from the Caltech\cite{caltaech} dataset on a neuromorphic sensor. While it includes face images, it does not provide keypoint annotations and contains only frontal views of subjects. The FES and NEFER datasets also offer event-based facial data, with FES using the Gen3 ATIS sensor\cite{ref-gen3atis} and NEFER employing the Gen4 CD sensor\cite{ref-gen4}. However, NEFER has not yet released annotations, and FES does not provide RGB frames. As a result, none of these datasets fully meet the need for synchronized RGB and event data with keypoint annotations. To overcome these limitations, we construct and utilize three distinct datasets. E-CelebV-HQ serves as our primary large-scale synthetic dataset for both model training and internal ablation studies. A comprehensive evaluation against state-of-the-art methods is performed on two challenging benchmarks, consisting of the synthetic E-WFLW-V dataset and our self-collected real-world E-SIE dataset.

\subsubsection{\textbf{E-CelebV-HQ}} We constructed the synthetic dataset E-CelebV-HQ to serve as our primary large-scale dataset for training and ablation studies. We employed the v2e\cite{v2e} event simulator with frame interpolation while setting the event threshold parameter to 0.2. The 35,664 event streams generated from the CelebV-HQ videos\cite{celebvhq} were segmented into 25 fps intervals, creating three event representations which are frame, voxel, and timesurface. From these representations, we observed that segments with minimal motion produced very few events. To mitigate this and ensure data quality, we selected the single frame with the highest event count from each video. We then used the corresponding RGB frames for annotation, placing 98 facial keypoints via SLPT\cite{slpt22} and using those as the ground-truth labels. Finally, this curated data was split into 28,531 samples for training, 713 for validation, and 6,420 for testing. The training set is used for both SSMER pre-training and CMFA fine-tuning, while the validation split is used only during CMFA-based alignment training. To obtain a high-confidence evaluation subset from the test split, we manually verified the pseudo labels and retained 1,554 images. The detailed construction diagram is shown in Fig.~\ref{fig:datapipe}.

\subsubsection{\textbf{E-WFLW-V}} For a standardized evaluation, we created E-WFLW-V, a synthetic event-based version of the public WFLW-V benchmark \cite{wflwv}. We converted 1,000 RGB clips from the original dataset into event streams using the same simulator settings as for E-CelebV-HQ. Each event stream was subdivided into segments corresponding to the original RGB frame rate. From these, the single segment with the highest event count was selected for each clip. This process yielded a final test set of 1,000 representative event images, each with 98 ground-truth landmark labels.

\subsubsection{\textbf{E-SIE}} Our self-collected event face dataset, E-SIE, comprises nine volunteers (aged 21 to 43, all of Asian descent) who continuously moved their heads. Illumination ranges from 30 to 120 lux among the three primary factors. Each subject participated in both eyeglasses-on and off conditions, with head speed either normal or fast. Head pose variation includes horizontal and vertical translations. These combined factors yield 8 distinct scenarios across 72 videos. By extracting 10 frames from each video, we obtained a total of 720 images exclusively used for evaluation.

\begin{table*}
  \centering
  \caption{Ablation of proposed components on the E-SIE dataset.}
  \resizebox{1\textwidth}{!}{
  \begin{tabular}{cccccccccccccc}
  \toprule
  \multirow{2}{*}{SSMER}& \multirow{2}{*}{CMFA}& \multicolumn{3}{c}{All} & \multicolumn{3}{c}{Speed (Fast)} & \multicolumn{3}{c}{Illumination (Low)} & \multicolumn{3}{c}{Eyeglasses} \\ \cmidrule(lr){3-5} \cmidrule(l){6-8} \cmidrule(l){9-11} \cmidrule(l){12-14} 
  & & NME& FR\textsubscript{10}& AUC\textsubscript{10}& NME& FR\textsubscript{10}& AUC\textsubscript{10}& NME& FR\textsubscript{10}& AUC\textsubscript{10}& NME& FR\textsubscript{10}& AUC\textsubscript{10} \\
  \midrule
     - & - & 14.720 & 57.306 & 0.183 
& 18.278& 67.111& 0.141& 15.167& 62.889& 0.150& 14.144& 54.944&0.196\\
 \checkmark& -& 14.801& 56.500& 0.185& 17.630& 65.111& 0.143& 15.419& 60.944& 0.155& 14.173& 54.889&0.194\\
     - & \checkmark & 8.596 & 28.083 & 0.369  & 10.675& 39.056& 0.297& 10.379& 39.722& 0.275& 8.633 & 28.056 &0.371 
\\
    \checkmark& \checkmark & \textbf{8.162} & \textbf{25.861} & \textbf{0.376}  & \textbf{9.776}& \textbf{34.500}& \textbf{0.321}& \textbf{9.832}& \textbf{37.389}& \textbf{0.278}& \textbf{8.004} & \textbf{24.556} & \textbf{0.390} 
\\
    \bottomrule
  \end{tabular}
  }
  \label{tab:ab_esie}
\end{table*}

\begin{figure*}
  \centering
   \includegraphics[width=0.95\linewidth]{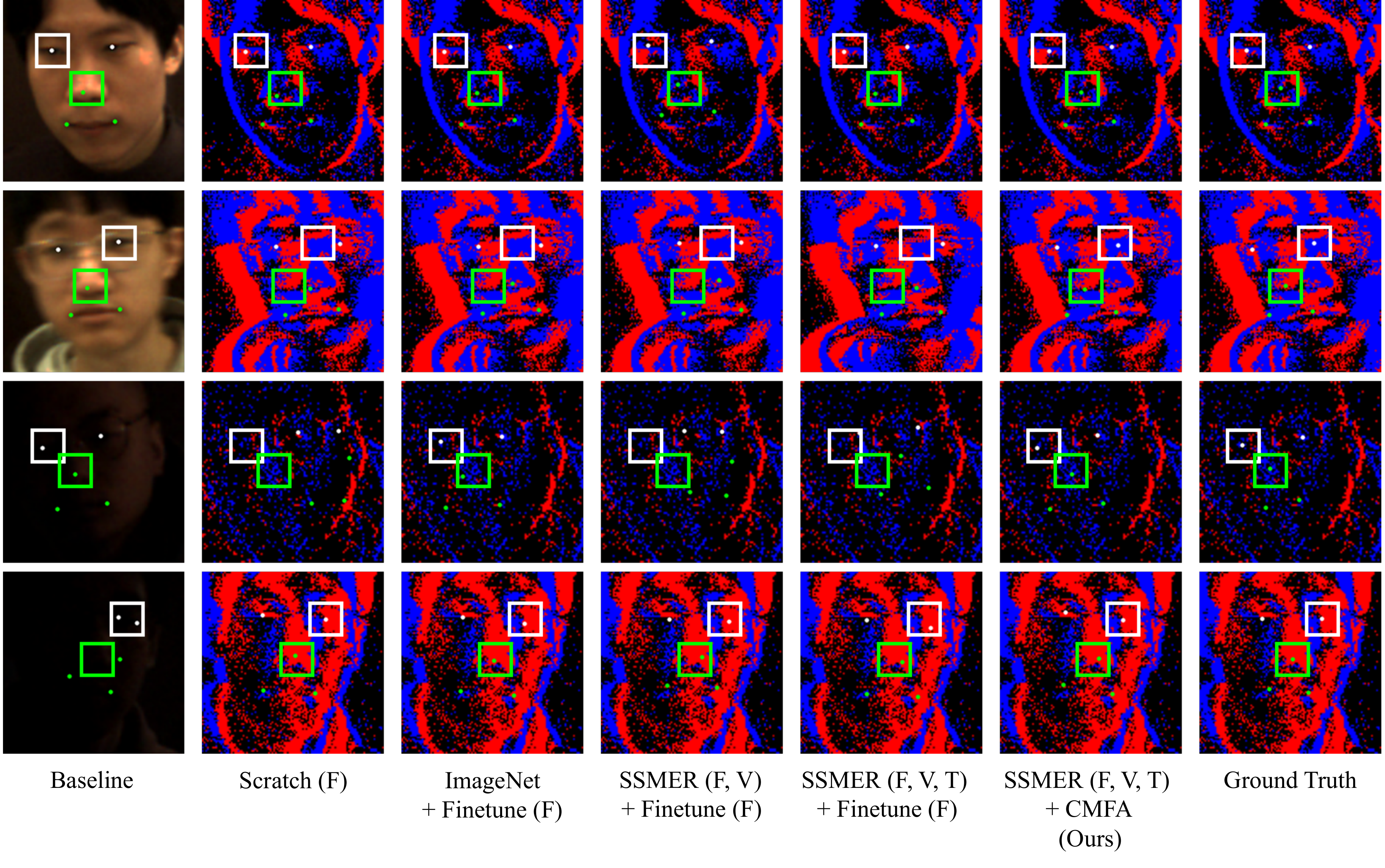}
   \caption{Ablation study visualization on our E-SIE dataset. Rows 1 to 4 represent the following conditions: normal, fast movement with eyeglasses, low illumination with eyeglasses, and low illumination with fast movement. F denotes Frame, V denotes Voxel, and T denotes Timesurface.}
   \label{fig:ssmer_ablation}
\end{figure*}

Our models are implemented in PyTorch and trained on an NVIDIA A6000 Ada GPU with 48GB of memory. For the SSMER-based event backbone, we employ SGD with a momentum of 0.9, a batch size of 128, a learning rate of 0.05, and train for 200 epochs. For the CMFA-based facial keypoint alignment fine-tuning, we use Adam with a batch size of 64, a learning rate of $1\times10^{-3}$, and train for 200 epochs. The loss function is a normalized L2 loss. We evaluate performance using normalized mean error (NME), failure rate (FR), and area under the curve (AUC), setting a threshold of 0.1 for FR and AUC. For NME normalization, E-WFLW-V uses the inter-ocular distance which is the distance between the outer corners of the eyes, while E-SIE uses the inter-pupil distance which is the distance between pupil centers. The detailed architecture and learning configuration of SSMER and CMFA are summarized in Table~\ref{tab:param}, which lists the module components, parameter counts, and training status (trainable or frozen). During SSMER pre-training, only the event backbone is optimized to learn event representations. During CMFA fine-tuning, the event backbone, CMFA module, and alignment head are updated, while the RGB backbone remains frozen. We utilize the RGB modality as an auxiliary guide that provides rich spatial information. Since the RGB backbone is already well-initialized with robust spatial features, updating it simultaneously with the event network might dilute these learned representations. Instead, by keeping the parameters frozen, we can leverage the abundant spatial information more effectively, ensuring it acts as a stable reference for cross-modal alignment. To illustrate the training and validation responses of our framework, we provide the learning curves in Fig.~\ref{fig:curve}. The left plot shows the step loss and average loss during SSMER pre-training, demonstrating stable convergence of the event backbone learning. The right plot presents the training loss and validation NME during CMFA fine-tuning, confirming consistent optimization behavior during cross-modal alignment training.

\subsection{Comparison with State-of-the-Art Method}
\label{sub:comparison}
\subsubsection{\textbf{Baselines}} We compare our method with seven state‑of‑the‑art facial‑landmark approaches. Because no existing model fuses RGB and event data for this task, we build a simple late‑fusion baseline: each RGB model is fine‑tuned on CelebV‑HQ, an event model is trained on E‑CelebV‑HQ, and their outputs are averaged to produce the final landmark prediction. In the comparison, the models AWing\cite{awing19}, Wing\cite{wing18}, STAR\cite{star23}, and ADNet\cite{adnet21} do not provide pretrained backbone weights. For this reason, we trained each model from scratch, following the experimental settings and hyperparameters described in the respective papers. In contrast, HRNet\cite{hrnet21}, SDFL\cite{sdfl21}, and PIPNet\cite{pipnet21} utilize backbone networks pretrained on ImageNet\cite{imagenet09}. For these models, we followed the training protocols outlined in the original papers and fine-tuned them on the E-CelebV-HQ dataset.

\subsubsection{\textbf{Results on E-SIE}} In Table~\ref{tab:res_esie_ewflwv}, we compare our model with previous methods on our real event face dataset, E-SIE, and tested on the event frame representation\cite{frame} with five keypoints. As shown in Table~\ref{tab:res_esie_ewflwv}, our method achieves state-of-the-art performance for NME, FR\textsubscript{10}, and AUC\textsubscript{10}. In particular, compared to an ADNet\cite{adnet21}, our model demonstrates a 28.6\% improvement in NME (reducing it from 11.434 to 8.162) and a 39.3\% improvement in FR\textsubscript{10} (from 42.611 to 25.861). Qualitative comparison results are provided in Fig. \ref{fig:comparison}, showing samples from E-SIE subsets under low light, fast movement, and eyeglasses-wearing conditions. We also evaluate a variant where the RGB backbone is updated during fine-tuning. As shown in Table~\ref{tab:res_esie_ewflwv}, this variant provides only marginal improvement in NME while resulting in worse FR\textsubscript{10} and AUC\textsubscript{10}.

\begin{table}[t]
  \centering
  \caption{Ablation of proposed components on the E-CelebV-HQ dataset.}
  \resizebox{0.7\columnwidth}{!}
  {  
  \begin{tabular}{ccccc}
      \toprule
SSMER& CMFA& NME & FR\textsubscript{10} &AUC\textsubscript{10}\\
  \midrule
    -& -& 3.136& 3.443& 0.704\\
 \checkmark & -& 3.290& 4.144&0.695\\
    -& \checkmark & 2.153& 1.491& 0.794\\
    \checkmark & \checkmark & \textbf{2.083}& \textbf{1.328}& \textbf{0.800}\\
    \bottomrule
  \end{tabular}
}
  \label{tab:ab_ecelebv_ts}
\end{table}

\begin{table}
  \centering
  \caption{Ablation for SSMER-based event backbone on the E-SIE dataset.}
  \resizebox{0.9\columnwidth}{!}
  {
  \begin{tabular}{lccc}
      \toprule
SSMER& NME & FR\textsubscript{10}& AUC\textsubscript{10}\\
  \midrule
    Frame& 9.266 & 32.056 & 0.337  \\
    Frame + Voxel& 8.709 & 28.583 & 0.365  \\
    Frame + Timesurface & 8.728 & 29.194 & 0.360  \\
    Frame + Voxel + Timesurface& \textbf{8.162} & \textbf{25.861} & \textbf{0.376}  \\
    \bottomrule
  \end{tabular}
    }
  \label{tab:ssmer_esie}
\end{table}

\subsubsection{\textbf{Results on E-WFLW-V}} We also compare our method with the previous approaches on our synthetic event face dataset, E-WFLW-V, evaluating all models on the event frame representation\cite{frame}, as shown in Table~\ref{tab:res_esie_ewflwv}. Since E-WFLW-V is a newly introduced dataset, no prior results exist for the baseline methods. Therefore, to ensure a fair comparison, we utilized the official open-source codes of all baseline models and re-trained or fine-tuned them on the corresponding datasets. Our method achieves the lowest NME, with a 9.0\% improvement (reducing from 5.200 to 4.731) over the best-performing baseline\cite{star23}. Although our FR\textsubscript{10} and AUC\textsubscript{10} do not surpass all baselines, they remain highly competitive.

\subsection{Ablation Study}
\label{sub:ablation}

\begin{table}[t]
  \caption{Ablation for SSMER-based event backbone on the E-CelebV-HQ dataset.}
  \centering
\resizebox{0.9\columnwidth}{!}{   
  \begin{tabular}{lccc}
      \toprule
SSMER& NME& FR\textsubscript{10}&AUC\textsubscript{10}\\
  \midrule
    -& 2.153& 1.491& 0.794\\
 Timesurface& 2.106& 1.391&0.797\\
    Timesurface + voxel& 2.102& 1.442& 0.799\\
    Timesurface + Frame& 2.095& 1.435& 0.799\\
    Timesurface + Frame + Voxel& \textbf{2.083}& \textbf{1.328}& \textbf{0.800}\\
    \bottomrule
  \end{tabular}
    }
  \label{tab:ab_ssmer_ts_ecelebv}
\end{table}

\subsubsection{\textbf{Ablation of proposed components}} To analyze the individual contributions of our core components, CMFA and SSMER, we conducted ablation studies on both the real-world E-SIE dataset and the synthetic E-CelebV-HQ dataset.
As shown in Table~\ref{tab:ab_esie}, we evaluated the components on E-SIE with the frame representation using both the entire dataset and three key variation subsets (fast speed, low illumination, and eyeglasses), keeping other conditions at their default settings. When only SSMER is applied, performance remains close to that of the baseline across all subsets. By contrast, introducing CMFA alone yields substantial gains, including a 41.6\% reduction in NME, a 51.0\% reduction in FR\textsubscript{10}, and a 101.6\% improvement in AUC\textsubscript{10} on the entire E-SIE dataset. Combining CMFA with SSMER provides further improvements over using CMFA alone, including a 5.0\% reduction in NME, a 7.9\% reduction in FR\textsubscript{10}, and a 1.9\% increase in AUC\textsubscript{10}. These findings show that SSMER is particularly beneficial when used together with CMFA.
To further validate our approach, we present additional ablation results on E-CelebV-HQ using the timesurface representation, as shown in Table~\ref{tab:ab_ecelebv_ts}. The effects of ablating SSMER and CMFA on this dataset align with those observed on E-SIE, confirming the robustness of our framework.

\subsubsection{\textbf{Effectiveness of Multi-Event Representation in SSMER}} We examined the impact of our SSMER-based event backbone, which integrates three event representations: frame\cite{frame}, voxel\cite{voxel18}, and timesurface\cite{ts17}. To compare different backbones, we trained each one with various event representations, then fine-tuned a CMFA-based keypoint alignment model using the event frame representation\cite{frame}. Our baseline is a SimSiam\cite{simsiam} model that relies solely on the frame representation. We also tested two pairwise combinations (frame and voxel, frame and timesurface) against our full SSMER approach, which merges all three. As shown in Table~\ref{tab:ssmer_esie}, the pairwise backbones already improve NME by up to 6.0\% over the single-representation baseline on E-SIE with the frame representation. Our final three-representation SSMER further boosts NME by 5.9\%. However, the improvement remains moderate because the frame, voxel, and timesurface representations are derived from the same event stream and therefore contain partially overlapping information. Due to the inherent sparsity of event data, the amount of complementary spatial information across these representations is limited. This observation is further supported by SSMER-specific ablations on E-CelebV-HQ with the timesurface representation, detailed in Table~\ref{tab:ab_ssmer_ts_ecelebv}.

\begin{figure}
  \centering
   \includegraphics[width=0.95\linewidth]{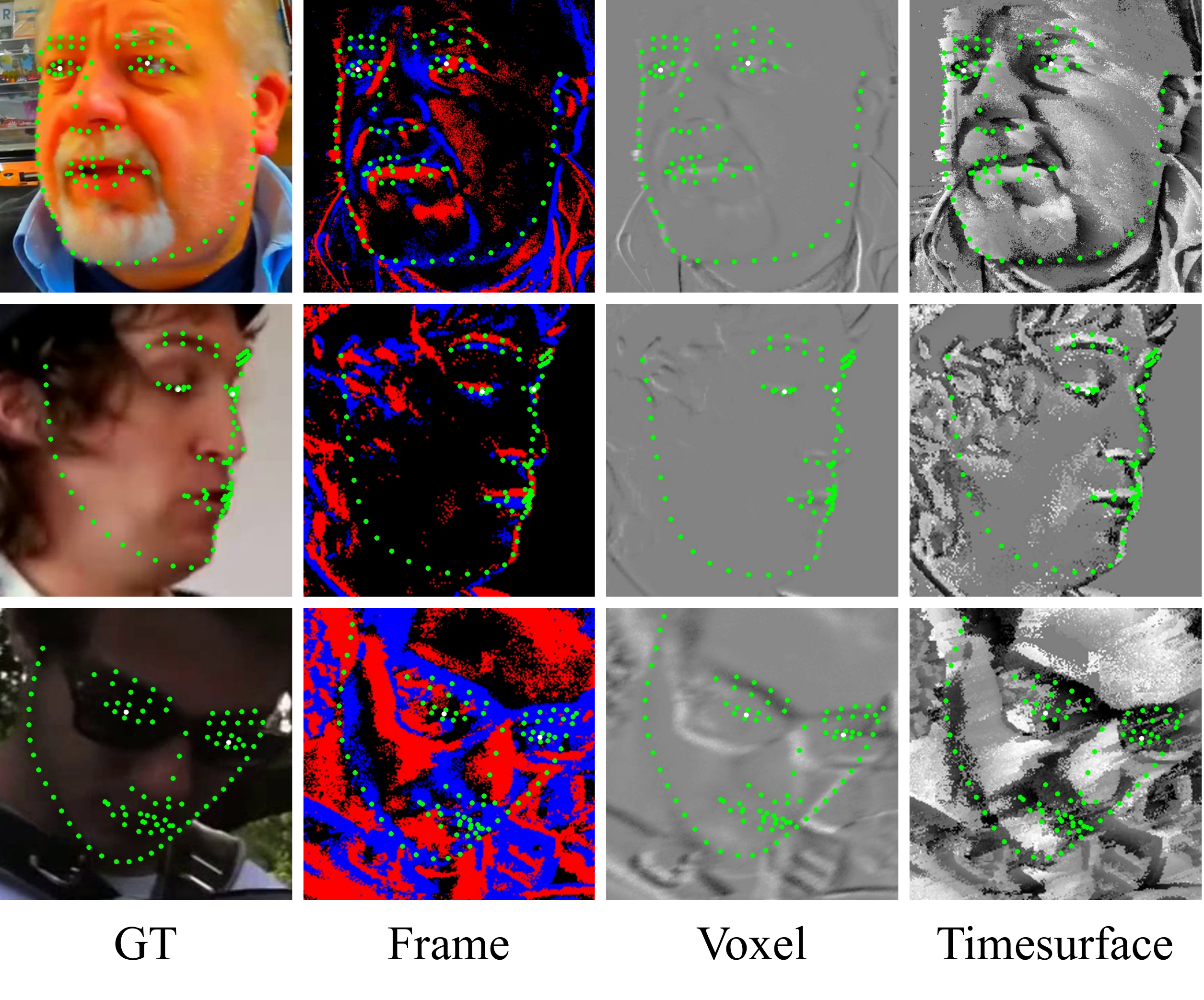}
   \caption{Visualization of ground truth and proposed method results using multiple event representations on E-WFLW-V.}
   \label{fig:ewflwv_multirepresentation}
\end{figure}

\begin{table*}
\caption{Performance comparison on event-based facial keypoint alignment, evaluated on E-SIE and E-WFLW-V\\(all models fine-tuned on E-WFLW-V).}
  \centering
    \resizebox{0.8\textwidth}{!}{
    \begin{tabular}{lccc ccc}
    \toprule
    \multirow{2}{*}{Models (RGB + Event)} & \multicolumn{3}{c}{E-SIE (Real)} & \multicolumn{3}{c}{E-WFLW-V (Synthetic)} \\
    \cmidrule(lr){2-4} \cmidrule(l){5-7}
    & NME(\%) $\downarrow$ & FR\textsubscript{10}(\%) $\downarrow$ & AUC\textsubscript{10} $\uparrow$ 
    & NME(\%) $\downarrow$ & FR\textsubscript{10}(\%) $\downarrow$ & AUC\textsubscript{10} $\uparrow$ \\
    \midrule
 HRNet\cite{hrnet21}& 27.966 & 61.333 & 0.157 
& 4.990 & 4.897 &0.789 
\\
 SDFL\cite{sdfl21}& 23.191 & 72.528 & 0.114 
& 10.736 & 32.298 &0.380 
\\
 Wing\cite{wing18}& 17.893 & 53.611 &0.206&    
3.104& 3.710 &0.729 
\\
 PIPNet\cite{pipnet21}& 14.932 & 58.639 & 0.158 
& 2.682 & 2.587 &0.750 
\\
 STAR\cite{star23}& 14.795 & 48.361 & 0.227 
& 2.723 & 2.826 &0.750 
\\
 AWing\cite{awing19}& 12.109 & 43.556 & 0.247 
& 3.361 & 3.916 &0.694 
\\
 ADNet\cite{adnet21}& 11.397 
& 41.417 & 0.261 & 3.441 & 4.996 &0.715 \\
 SLPT\cite{slpt22}& 8.365 & 29.667 & 0.334 
& 2.397 & 2.217 &0.775 
\\
    \midrule
    \textbf{Ours} & \textbf{6.431}& \textbf{16.694}&\textbf{0.442}&    \textbf{2.156}& \textbf{1.501}&\textbf{0.791}\\
    \bottomrule
  \end{tabular}
}
  \label{tab:comp_wflwv}
\end{table*}

Fig. \ref{fig:ssmer_ablation} provides a qualitative visualization of these ablation studies on the E-SIE dataset. It presents our SSMER + CMFA approach against various baselines. These include learning from scratch with event frames, fine-tuning an ImageNet-pretrained backbone, and fine-tuning SSMER backbones trained with different event pairs, such as (frame~\cite{frame}, voxel~\cite{voxel18}) and (frame~\cite{frame}, voxel~\cite{voxel18}, timesurface~\cite{ts17}). For reference, results from an RGB-only baseline~\cite{slpt22} are also included. The visualizations confirm that our method is the only one to succeed across all four challenging scenarios, while the RGB-based method fails under low illumination and fast movement. These results visually reaffirm the quantitative findings: SSMER alone offers limited improvement, but combining it with CMFA leads to a substantial performance boost.

\subsubsection{\textbf{Analysis on Additional Fine-tuning}}
To assess the benefit of training on more data with high-motion characteristics, we conducted an additional fine-tuning experiment. For this purpose, we created an expanded training set from the E-WFLW-V clips. Specifically, we extracted the 10 segments with the highest event counts from each of the 1,000 clips, resulting in a larger dataset of 10,000 event images. We then split this dataset into 8,000 images for training and 2,000 for testing and proceeded to fine-tune our model. This fine-tuning setup allows us to assess the benefit of additional event-based training data under high-motion conditions. The results of this experiment are reported in Table~\ref{tab:comp_wflwv}.

\begin{figure}
  \centering
   \includegraphics[width=0.95\linewidth]{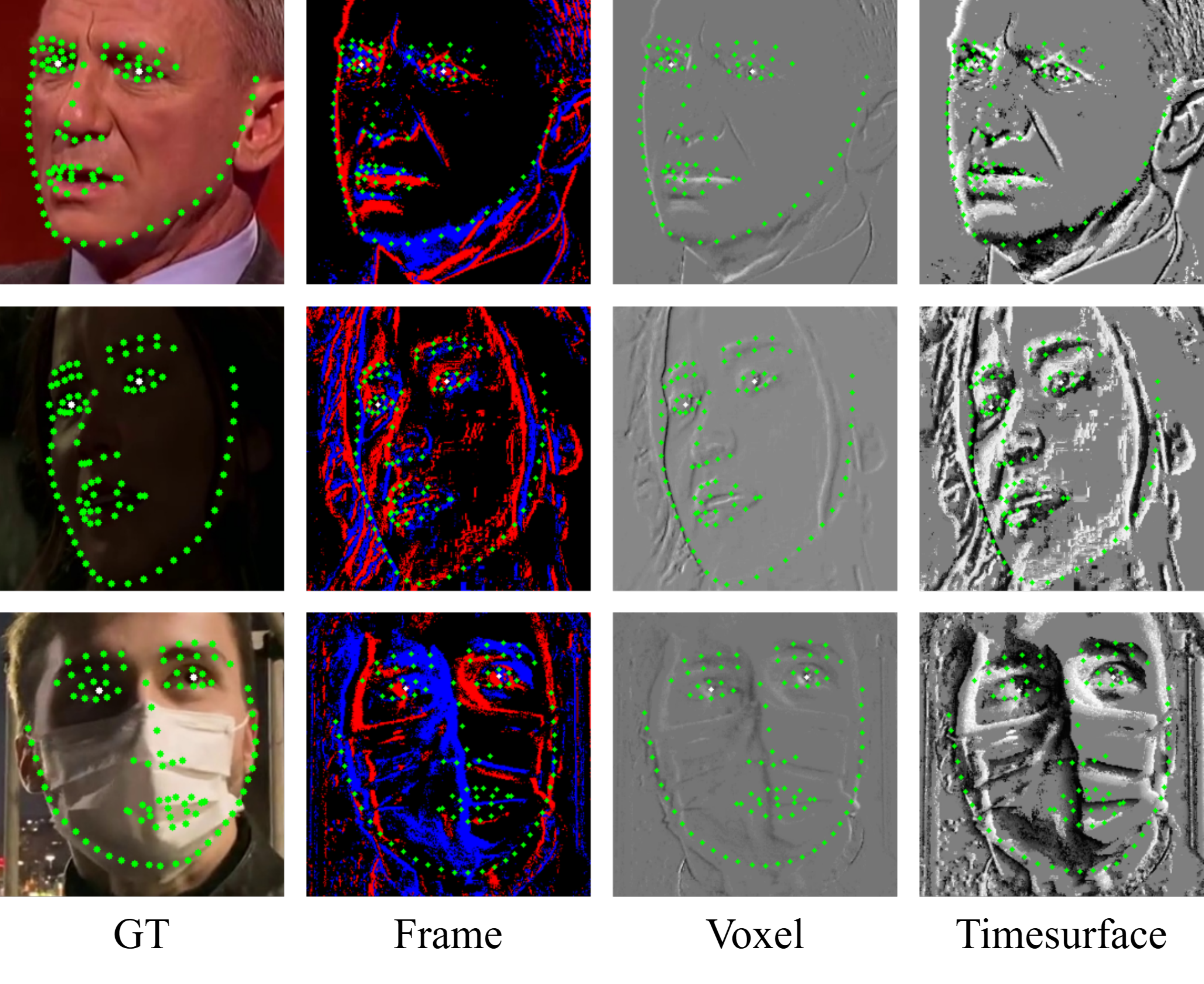}
   \caption{Visualization of ground truth and proposed method results using multiple event representations on E-CelebV-HQ.}
   \label{fig:ecelebvhq_multirepresentation}
\end{figure}

\subsection{Discussion and Future Work} Our approach, built on SSMER and CMFA, achieves robust event-based facial keypoint alignment. Additional event representations or integrating CMFA with other RGB-based methods may offer further enhancements. In particular, E-SIE yields a higher NME than E-WFLW-V, as E-WFLW-V includes 98 keypoints, which leads to a lower NME than E-SIE’s 5 keypoints (eyes, nose, mouth). This gap reflects the challenge of aligning finer facial details in event data with limited spatial information. Fig. \ref{fig:ewflwv_multirepresentation} presents E-WFLW-V evaluation results across various event representations, and Fig. \ref{fig:ecelebvhq_multirepresentation} shows the corresponding E-CelebV-HQ results.} The NME differences across synthetic sets reflect a trade-off between scale and label quality: E-CelebV-HQ is large but pseudo-labeled, whereas E-WFLW-V provides manually refined landmarks, and our method still generalizes well to the unseen real-event E-SIE dataset. To further analyze robustness across different conditions, we provide a detailed comparison under diverse scenarios of the E-SIE dataset, including high illumination and normal motion, as shown in Table~\ref{tab:supp_esie}. The results indicate that our method achieves the best performance across all conditions, confirming that the proposed framework remains effective not only in challenging environments but also under normal illumination settings. The proposed pipeline is fine-tuned on E-CelebV-HQ, a synthetic dataset generated using the v2e\cite{v2e} simulator and supervised with pseudo-labels. As next steps, we will focus on higher-fidelity simulation and improved weakly supervised labeling to boost alignment accuracy. Moreover, our method does not rely on a specific sensor and can be applied across different event-camera models, supporting broader practicality as event hardware evolves. On the E-SIE dataset, the ResNet-based SSMER achieves NME 9.698\%, FR\textsubscript{10} 34.833\%, and AUC\textsubscript{10} 0.315, which is slightly lower than the HRNet-based SSMER. This result indicates that SSMER is not limited to HRNet and can also be applied to other backbone architectures, although HRNet provides better performance in our framework. Evaluating additional lightweight backbones remains an important direction for future work.

\begin{table*}
  \centering
  \caption{Detailed performance Comparison on E-SIE}
  \resizebox{0.9\textwidth}{!}{
  \begin{tabular}{cccccccccc}
  \toprule
  \multirow{2}{*}{Models (RGB + Event)} 
& \multicolumn{3}{c}{Speed (Normal)} & \multicolumn{3}{c}{Illumination (High)} & \multicolumn{3}{c}{No Eyeglasses}  \\ \cmidrule(r){2-4} \cmidrule(lr){5-7} \cmidrule(l){8-10} 
& NME& FR\textsubscript{10}& AUC\textsubscript{10}& NME& FR\textsubscript{10}& AUC\textsubscript{10}& NME& FR\textsubscript{10}& AUC\textsubscript{10}\\
  \midrule
     HRNet\cite{hrnet21}
& 27.483 & 93.944 & 0.016 & 24.730 & 94.222 & 0.013 
& 26.881 & 92.778 & 0.021 
\\
 SDFL\cite{sdfl21}
& 27.483 & 93.944 & 0.016 
& 17.088& 66.278& 0.132& 26.881 & 92.778 & 0.021 
\\
 AWing\cite{awing19}
& 14.664 & 48.666 & 0.212 
& 13.461& 44.555 & 0.235 & 13.589 & 51.222 & 0.196 
\\
 Wing\cite{wing18}
& 11.488 & 39.278 & 0.274 
& 11.100& 35.333 & 0.298 & 11.710 & 43.556 & 0.242 
\\
 PIPNet\cite{pipnet21}
& 11.424 & 55.445 & 0.136 
& 11.472& 53.223 & 0.138 & 12.505 & 61.112 & 0.117 
\\
 STAR\cite{star23}
& 10.434 & 37.833 & 0.274 
& 9.642& 33.388 & 0.304 & 11.695 & 45.499 & 0.233 
\\
 ADNet\cite{adnet21}
& 9.724 & 35.666 & 0.289 & 9.468& 31.944 & 0.315 & 11.119 & 42.055 & 0.246 
\\
    \midrule
 \textbf{Ours} & \textbf{6.548}& \textbf{17.222}& \textbf{0.431}& \textbf{6.492}& \textbf{14.333}& \textbf{0.474}& \textbf{8.32}& \textbf{27.166}& \textbf{0.362}
\\ \midrule
    
 \multirow{2}{*}{Models (RGB + Event)} 
& \multicolumn{3}{c}{Speed (Fast)}& \multicolumn{3}{c}{Illumination (Low)}& \multicolumn{3}{c}{Eyeglasses}\\
 \cmidrule(r){2-4} \cmidrule(lr){5-7} \cmidrule(l){8-10} 
& NME& FR\textsubscript{10}& AUC\textsubscript{10}& NME& FR\textsubscript{10}& AUC\textsubscript{10}& NME& FR\textsubscript{10}&AUC\textsubscript{10}\\ \midrule
 HRNet\cite{hrnet21}& 28.701 & 93.778 & 0.016 
& 31.454 & 93.500 & 0.019 
& 29.303 & 94.944 &0.011 
\\
 SDFL\cite{sdfl21}
& 22.238 & 76.500 & 0.089 
& 25.520 & 82.278 & 0.068 
& 22.001 & 74.778 &0.106 
\\
 AWing\cite{awing19}
& 14.844 & 57.056 & 0.174 
& 16.047 & 61.167 & 0.151 
& 15.919 & 54.500 &0.190 
\\
 Wing\cite{wing18}
& 13.724 & 52.722 & 0.202 
& 14.112 & 56.667 & 0.178 
& 13.502 & 48.444 &0.234 
\\
 PIPNet\cite{pipnet21}
& 13.552 & 66.111 & 0.100 
& 13.504 & 68.333 & 0.098 
& 12.471 & 60.444 &0.119 
\\
 STAR\cite{star23}
& 13.152 & 52.333 & 0.198 
& 13.944 & 56.778 & 0.168 
& 11.891 & 44.667 &0.239 
\\
 ADNet\cite{adnet21}
& 13.144 & 49.556 & 0.211 
& 13.400 & 53.278 & 0.185 
& 11.749 & 43.167 &0.254 
\\ \midrule
 \textbf{Ours} & \textbf{9.776}& \textbf{34.500}& \textbf{0.321}& \textbf{9.832}& \textbf{37.389}& \textbf{0.278}& \textbf{8.004}& \textbf{24.556}&\textbf{0.390}\\ \bottomrule
  \end{tabular}
  }
  \label{tab:supp_esie}
\end{table*}

Table~\ref{tab:nofusion_esie} presents the performance of the event-only configuration to assess the impact of cross-modal fusion. On E-SIE, the event-only setting yields NME 14.801, FR 56.5\%, and AUC 0.185, whereas RGB and event yields NME 8.162, FR 25.86\%, and AUC 0.376. This shows that CMFA is key to compensating for the limited spatial detail of events. This also reflects the sensing gap between frame-based RGB methods and event-based inputs, where the absence of texture and appearance cues poses challenges for RGB-optimized landmark detectors, particularly under low light or fast motion. As future work, we will develop an event-only variant that removes dependence on RGB. Our current design uses RGB as an auxiliary guide, which can be weak in low light, HDR, or fast motion. As future research, we will add an event-only mode and a confidence-aware fusion that reduces RGB weight and increases event weight in difficult situations. While E-SIE was collected in a controlled setting to systematically vary speed, illumination, and eyeglasses, it cannot cover all in-the-wild conditions. We plan to construct a larger, unconstrained event-face dataset to further assess robustness. Additionally, we plan to extend our framework to address the sparsity challenge in far-distance scenarios (beyond 1 m) and validate the real-time feasibility of our method on embedded systems for practical deployment. 

Table~\ref{tab:computational_cost} reports complexity and runtime. All FLOPs, parameters, and FPS reported in Table~\ref{tab:computational_cost} are measured by us under the same hardware environment and input size. For fair comparison, RGB-only baselines are adapted to process both RGB and event inputs using a late-fusion protocol. Our model achieves 9.15 GFLOPs, 21.62 M parameters, and the highest throughput of 17.865 FPS. RGB and event feature aggregation via CMFA introduces additional computational overhead, which may limit deployment on resource-constrained embedded platforms. Future work will focus on lightweight architectures for edge-oriented real-time inference.

\begin{table}
\caption{Performance comparison on event-based facial keypoint alignment, evaluated on E-SIE (Event only)}
  \centering
    \resizebox{1\linewidth}{!}{
  \begin{tabular}{lccc}
    \toprule
 \multirow{2}{*}{Models (Event only)}& \multicolumn{3}{c}{E-SIE (Real)} \\ 
 \cmidrule(lr){2-4}
 & NME(\%) $\downarrow$ & FR\textsubscript{10}(\%) $\downarrow$ &AUC\textsubscript{10} $\uparrow$ 
\\
    \midrule
    HRNet\cite{hrnet21}&     45.022& 68.583 &0.123 
\\
 AWing\cite{awing19}& 30.438 & 83.444 &0.060 
\\
 SDFL\cite{sdfl21}& 30.407 & 79.917 &0.078 
\\
 Wing\cite{wing18}&  22.533 & 67.222 &0.130 
\\
 ADNet\cite{adnet21}&  21.617 & 65.194 &0.140 
\\
 STAR\cite{star23}&  20.910 & 65.778 &0.141 
\\
 PIPNet\cite{pipnet21}&  16.628 & 89.827 &0.102 
\\
    \midrule
    \textbf{Ours (Event only)} & \textbf{14.801}& \textbf{56.500}&\textbf{0.185}\\
    \bottomrule
  \end{tabular}
}
\label{tab:nofusion_esie}
\end{table}

\begin{table}
\caption{Comparative analysis of model complexity (measured by us under a unified RGB and event input setting)}
  \centering
  \resizebox{1\linewidth}{!}{
  \begin{tabular}{lccc} 
  \toprule
  \makecell[l]{Models \\ (RGB + Event)} & FLOPs(G) ($\downarrow$)& Params(M) ($\downarrow$) & FPS ($\uparrow$) \\
  \midrule
  SDFL\cite{sdfl21}  & \textbf{8.92} & 22.18 & 11.667 \\
  AWing\cite{awing19} & 26.952& 34.974& 12.173 \\
  Wing\cite{wing18}  & 26.952& 34.974& 12.445 \\
  STAR\cite{star23}  & 26.952& 34.974& 12.557 \\
  \midrule
  \textbf{Ours}      & 9.15 & \textbf{21.62}& 17.865\\
  \bottomrule
  \end{tabular}
}
\label{tab:computational_cost}
\end{table}

\section{Conclusion}
\label{sec:conclusion}
We present a novel framework for event-based facial keypoint alignment, unifying cross-modal fusion attention and self-supervised multi-event representation learning. Through extensive experiments on our self-collected real event dataset E-SIE and the synthetic E-WFLW-V dataset, our approach demonstrates stable alignment under challenging conditions like low light, rapid movement, and eyeglasses. The effectiveness of our modules is validated by significant improvements in accuracy and stability, and by ablation studies confirming that integrating CMFA and SSMER outperforms conventional baselines.


\begin{IEEEbiography}[{\includegraphics[width=1in,height=1.25in,clip,keepaspectratio]{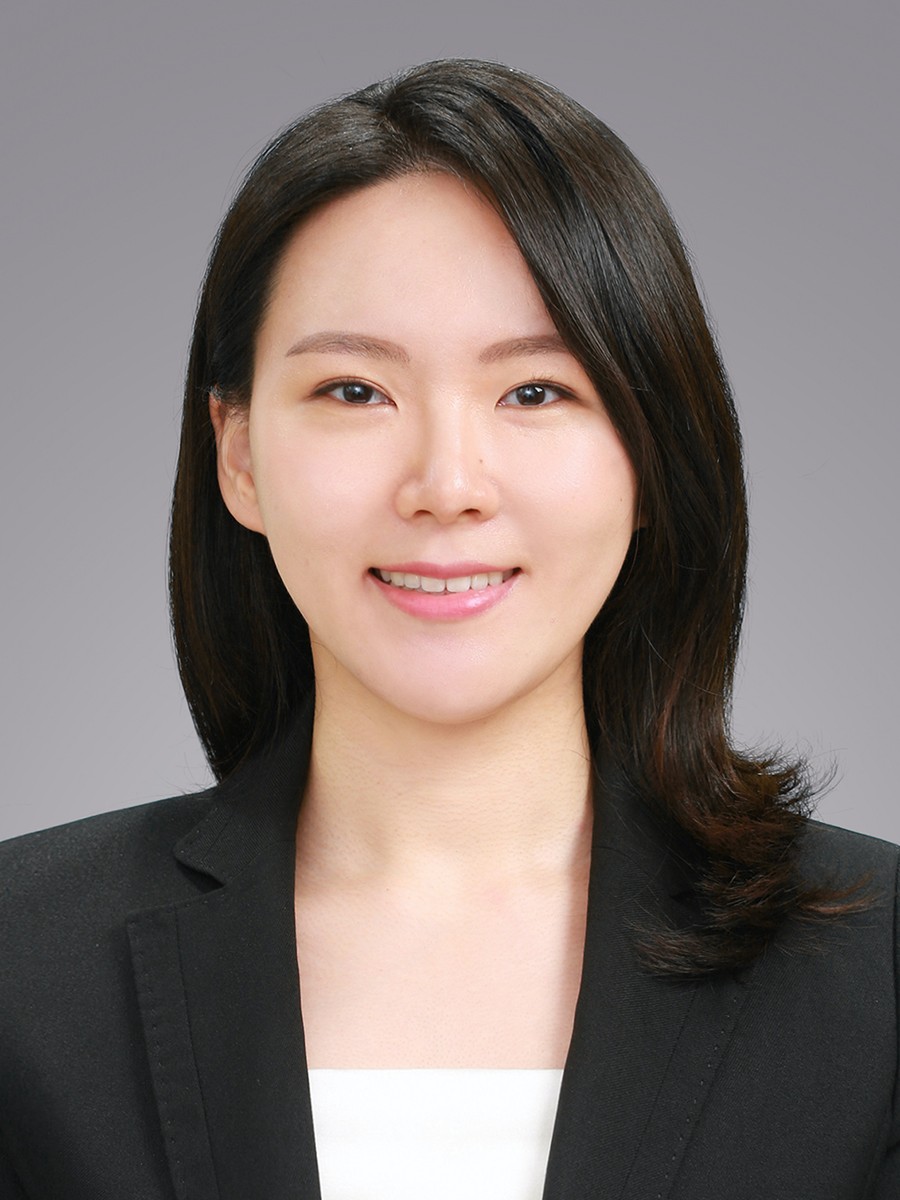}}]{Donghwa Kang} received the B.S. degree in electronic and electrical engineering from Hongik University, Seoul, South Korea, in 2024, where she is currently pursuing the M.S. degree. Her research interests include computer vision, particularly in the areas of facial analysis, eye tracking, and event cameras.
\end{IEEEbiography}

\begin{IEEEbiography}[{\includegraphics[width=1in,height=1.25in,clip,keepaspectratio]{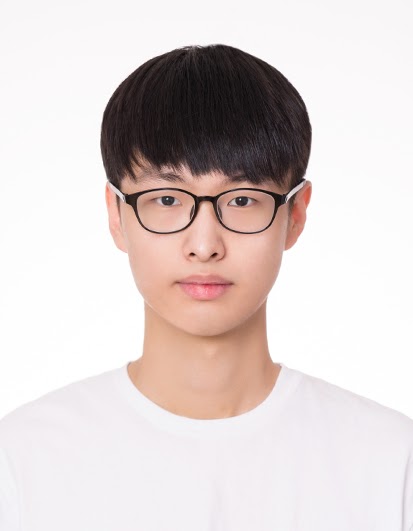}}]{Junho Kim} is currently pursuing the B.S. degree in electronic and electrical engineering at Hongik University, Seoul, South Korea. His research interests include computer vision, with a focus on inpainting, eye tracking, and self-supervised learning.
\end{IEEEbiography}

\begin{IEEEbiography}[{\includegraphics[width=1in,height=1.25in,clip,keepaspectratio]{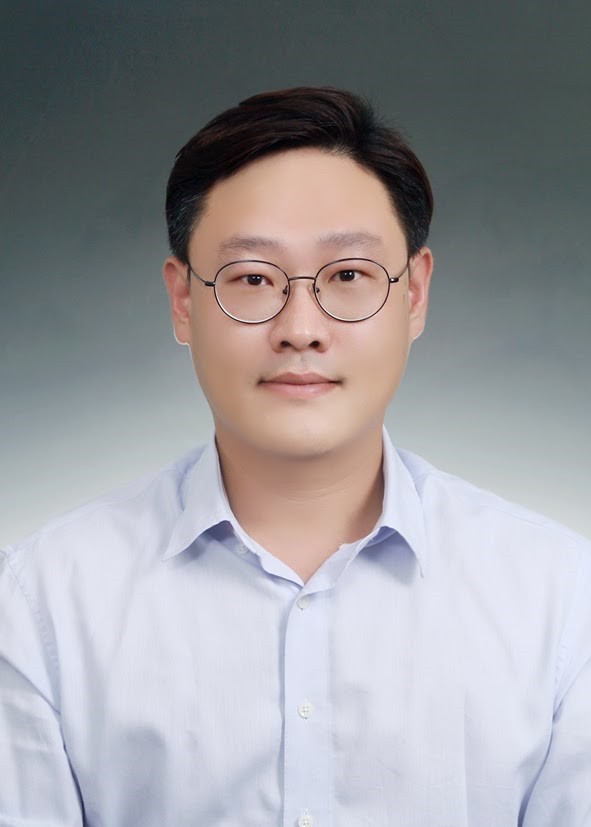}}]{Dongwoo Kang} received the B.S. degree in electrical engineering from Seoul National University, Seoul, South Korea, in 2007, and the M.S. and Ph.D. degrees in electrical engineering from the University of Southern California, Los Angeles, CA, USA, in 2009 and 2013, respectively. He was a Senior Researcher at Samsung Advanced Institute of Technology, Suwon, South Korea, from 2013 to 2021. In 2021, he joined the Faculty of the Department of Electronic and Electrical Engineering, at Hongik University, Seoul, South Korea, where he is currently an Assistant Professor. His research interests include the area of image processing and computer vision including detection, tracking, segmentation, image enhancement, application to augmented reality, autostereoscopic 3D displays, medical image analysis, and event cameras.
\end{IEEEbiography}
\end{document}